\documentclass[journal]{IEEEtran}
\usepackage{cite}

\usepackage{graphicx}
\ifCLASSINFOpdf
\else
\fi
\usepackage{amsmath}
\usepackage{array}

\usepackage{stfloats}
\usepackage{url}

\usepackage{multirow}
\usepackage{xr-hyper}
\usepackage[bookmarksopen,bookmarksdepth=2,breaklinks=true]{hyperref}
\usepackage{makecell}

\makeatletter
\let\MYcaption\@makecaption
\makeatother

\usepackage[font=footnotesize]{subcaption}

\makeatletter
\let\@makecaption\MYcaption
\makeatother

\usepackage{tabularx}
\usepackage[flushleft]{threeparttable}
\usepackage{booktabs}
\usepackage{empheq}
\newcolumntype{C}[1]{>{\centering\let\newline\\\arraybackslash\hspace{0pt}}m{#1}}
\usepackage{titlesec}
\titlespacing\section{0pt}{10pt plus 4pt minus 2pt}{2pt plus 2pt minus 2pt}
\titlespacing\subsection{0pt}{9pt plus 4pt minus 2pt}{1pt plus 2pt minus 2pt}
\titlespacing\subsubsection{0pt}{8pt plus 4pt minus 2pt}{0pt plus 2pt minus 2pt}

\expandafter\def\expandafter\normalsize\expandafter{%
    \normalsize
    \setlength\abovedisplayskip{5pt}
    \setlength\belowdisplayskip{5pt}
    \setlength\abovedisplayshortskip{4pt}
    \setlength\belowdisplayshortskip{4pt}
}

\begin{document}
%
\title{Design and Actuator Optimization of Lightweight and Compliant Knee Exoskeleton for Mobility Assistance of Children with Crouch Gait}
%
%
%

\author{Sainan~Zhang,
        Tzu-Hao~Huang,
        Chunhai~Jiao,
        Mhairi~MacLean,
        Junxi~Zhu,
        Shuangyue~Yu,
        Hao~Su$^\dag$,~\IEEEmembership{Member,~IEEE}%
        \thanks{This work is supported by the National Science Foundation National Robotics Initiative (NRI) grant IIS 1830613, NSF CAREER award CMMI 1944655. Any opinions, findings, and conclusions or recommendations expressed in this material are those of the authors and do not necessarily reflect the views of the funding organizations.}%
        \thanks{All authors are with Lab of Biomechatronics and Intelligent Robotics (BIRO), Department of Mechanical Engineering, The City University of New York, City College, NY, 10023, US.}%
        \thanks{$^\dag$ Corresponding author. (E-mail: hao.su@ccny.cuny.edu)}%
}

\maketitle

\begin{abstract}
Pediatric exoskeletons offer great promise to increase mobility for children with crouch gait caused by cerebral palsy. A lightweight, compliant and user-specific actuator is critical for maximizing the benefits of an exoskeleton to users. To date, pediatric exoskeletons generally use the same actuators as adult exoskeletons, which are heavy and resistive to natural movement. There is yet no easy way for robotic exoskeletons to accommodate the changes in design requirements that occur as a child ages. We developed a lightweight (1.65 kg unilateral mass) and compliant pediatric knee exoskeleton with a bandwidth of 22.6 Hz that can provide torque assistance to children with crouch gait using high torque density motor. Experimental results demonstrated that the robot exhibited low mechanical impedance (1.79 Nm average backdrive torque) under the unpowered condition and 0.32 Nm with zero-torque tracking control. Root mean square (RMS) error of torque tracking result is less than 0.73 Nm (5.7\% with respect to 12 Nm torque). To achieve optimal age-specific performance, we proposed the first optimization framework that considered both motor and transmission of the actuator system that can produce optimal settings for children between 3 and 18 years old. The optimization generated an optimal motor air gap radius that monotonically increases with age from 0.011 to 0.033 meters, and optimal gear ratio varies from 2.6 to 11.6 (3-13 years old) and 11.6 to 10.2 (13-18 years old), leading to actuators of minimal mass.
\end{abstract}

\begin{IEEEkeywords}
Pediatric knee exoskeleton, actuator optimization, crouch gait, compliant actuator.
\end{IEEEkeywords}

%
\IEEEpeerreviewmaketitle

\section{Introduction}
%
%
%
%
\IEEEPARstart{C}{}erebral palsy is a neuromuscular disorder that limits mobility and reduces the quality of life. Gait pathologies arising from cerebral palsy cause inflated metabolic cost and decreased walking speed \cite{carcreff2020walking}. The most common pathological gait due to cerebral palsy is crouch gait \cite{wren2005prevalence}, which is characterized by excessive flexion of the knee. Over 60\% of children with moderate-severe mobility impairment (gross motor function classification system (GMFCS) of II to IV) and over 45\% of children with mild impairment (GMFCS of I) exhibit increased knee flexion \cite{rethlefsen2017prevalence}. Children who have better walking ability at a young age are more likely to retain the ability to walk in adulthood \cite{palisano2010probability}. It is therefore imperative that children with cerebral palsy receive effective treatment at a young age \cite{bjornson2007ambulatory}. Treatments for crouch gait, which include physical therapy, orthotic devices, chemical injections, and surgery \cite{lerner2017lower}, may improve mobility or posture in the short term, but even with treatment, half of the children with cerebral palsy lose their ability to walk by the time they reach adulthood \cite{bottos2003ambulatory}.
\begin{figure}[!htbp]
  \centering
  \begin{subfigure}{0.95\columnwidth}
  \centering
    \includegraphics[width=\columnwidth]{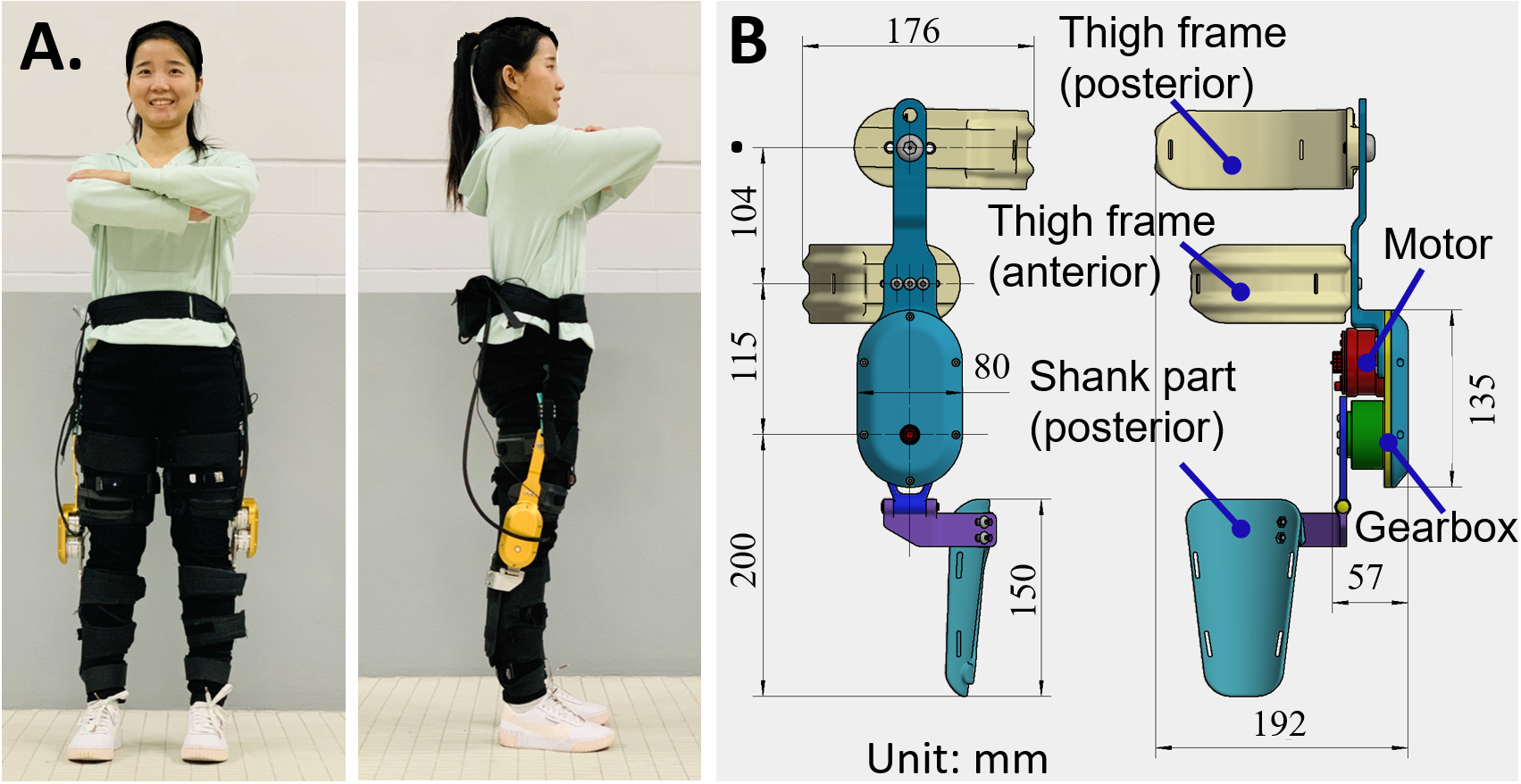}
    \phantomsubcaption
    \label{fig:Exoskeleton_Overview_Person}
  \end{subfigure}
  
  \vspace{-0.5cm}
  \begin{minipage}{\columnwidth}
    \phantomsubcaption
    \label{fig:Exoskeleton_Overview_Dimensions}
  \end{minipage}
  
  \caption{Overview of a portable knee exoskeleton designed for pediatric gait assistance in children with cerebral  palsy. (A) The front and side view of device.  A unilateral (bilateral) pediatric knee exoskeleton weighs $1.65$ ($2.78$) kg in total, including control electronics and battery ($0.7$ kg). (B) The front view and side view of mechanical design.}
  \label{fig:Exoskeleton_Overview}
\end{figure}

Robotic exoskeletons are emerging as a promising therapy for crouch gait. Exoskeletons can immediately improve gait outcomes like knee flexion and have no side effects (unlike drugs) \cite{lerner2017lower}. For example, the pediatric knee exoskeleton from the National Institutes of Health (NIH) improved knee extension in patients with gross motor function classification system (GMFCS) levels I and II (i.e. children who are ambulatory but need assistive devices to support ambulation) \cite{lerner2017lower}. However, the mainstream pediatric exoskeletons cannot be used for at-home therapy or to assist with daily-living because they are tethered systems, limiting their use to the clinic \cite{michmizos2015robot}. 
Past research found that entirely wearable exoskeletons would be more advantageous for rehabilitation than partially grounded exoskeletons \cite{gassert2018rehabilitation}. Therefore, an exoskeleton to assist daily living should be portable, completely wearable, and not restrict the range of motion.

To improve neurorehabilitation, an exoskeleton for individuals with crouch gait must promote active engagement and human efforts \cite{lerner2017lower}, in addition to improving walking ability. High compliance and bandwidth are crucial to encouraging active engagement, as compliance (low impedance) ensures the exoskeleton can be moved by the user and high bandwidth allows the exoskeleton to quickly respond to the user. Additionally, a heavy device will add a penalty to the energetic cost and alter inertial properties of the limbs, which could supersede any benefits provided by robotic actuation. Furthermore, user engagement and effort are discouraged by bulky and heavy devices. Exoskeletons are often limited by their weight \cite{sanz2017result, sawicki2020exoskeleton}, and state-of-the-art knee exoskeletons report high stiffness (low compliance) \cite{shepherd2017design}.

The mechatronic design of a pediatric exoskeleton is more challenging than that of an exoskeleton for adults. First, biomechanics parameters such as mass, joint torque, and joint velocity vary dramatically with age \cite{van2010age}. Additionally, a wide range of temporal-spatial, kinematic, and kinetic parameters can characterize crouch gait \cite{o2020crouch}, and the underlying neuromechanical causes of crouch gait can differ between individuals \cite{hicks2008crouched}. The wide range of gait parameters and biomechanical properties exhibited by children with crouch gait can make the design of an assistive exoskeleton challenging. Furthermore, the energetic penalty of walking increases proportionally with additional mass as a percentage of body weight \cite{grabowski2005independent}. Using the square-cube law, a child who is half the height of an adult will be about $1/8$ of their weight. However, the proportional difference in weight between adult and pediatric exoskeletons does not match the reduction in body mass \cite{lerner2017lower}. Therefore, the weight of pediatric exoskeletons as a percentage of body weight is much higher than in adults and imposes a greater percentage increase in the energetic cost of walking. In addition to high compliance and bandwidth, a portable, assistive pediatric exoskeleton must also be as lightweight as possible. 

User-specific optimized actuator design is critical to maximizing the dynamics of the human-robot interaction. As suggested above, a “one-size-fits-all” school of thought is not suitable for pediatric exoskeletons. Overpowered actuators unnecessarily increase weight, while underpowered actuators cannot provide effective gait assistance \cite{rossi2013feasibility}. There has been recent interest in transmission optimization for actuators in wearable robotics including optimization of the elastic component for series elastic actuation \cite{nieto2019minimizing} and gear train transmission \cite{bartlett2018design}. To the best of our knowledge, there has been no research on motor optimization for actuator design in wearable robotics. 

The purpose of this paper was to develop a lightweight, compliant exoskeleton to improve walking functionality in children with crouch gait, and further refine this design across a pediatric age range with an actuator optimization framework. The contribution of this paper includes: 1) mechatronic design of a fully portable, lightweight, and compliant knee exoskeleton for mobility assistance of children with crouch gait in community settings. 2) a generic optimization framework to customize both the electric motor and transmission design of wearable robots subject to user-specific and age-specific design requirements, including output torque, speed, bandwidth, and backdrive torque as constraints.

\section{Mechatronic Design and Control}
To design lightweight, compact, and compliant wearable robots for pediatric gait assistance in community settings, we customized high-torque density electric motors developed in our lab to meet pediatric needs and achieve the design of a compact robotic exoskeleton mechanism with compliant actuators.

\subsection{Mechanical Design}
The exoskeleton is composed of an actuator per leg, support frames for the thigh and shank, and a waist-mount system, as shown in \figurename~\ref{fig:Exoskeleton_Overview}. The actuator (see Section \ref{sec:High_Torque_Density_Motor}) transfers assistive torque through the support frames to the user’s knee joint. The support frames each fasten to the corresponding limb segment through a cuff with adjustable Velcro straps. We designed the support frame cuffs to comfortably fit the user by using 3D scans of the limb to define the geometry. We 3D printed these cuffs using carbon-fiber reinforced nylon. A double-hinge mechanism \cite{wang2018comfort} connects the actuator to the shank support frame, which provides a passive frontal plane degree of freedom. This hinge allows the device to passively align with user’s limb, safely transferring assistive torque while preventing unwanted parasitic loads. The thigh support frame is additionally supported by a waist-mount system, which consists of an elastic strap in tension between the thigh support frame and a waist belt. This support prevents the exoskeleton from slipping down the leg and avoid misalignment with the knee joint. For bilateral assistance, we connect the other actuator and support frame assembly to the waist-mount system.

The knee joint actuator is composed of a custom high torque density actuator, torque sensor, and embedded system for low-level control. The custom torque sensor transfers torque loads from the actuator output stage to the shank support frame through bolted connections on each end. The sensor is rated for torques up to $\pm\,40$ Nm with a resolution of $0.1$ Nm.

\vspace{-0.1cm}
\begin{figure}[hbt!]
    \centering
    \includegraphics[width=0.95\columnwidth]{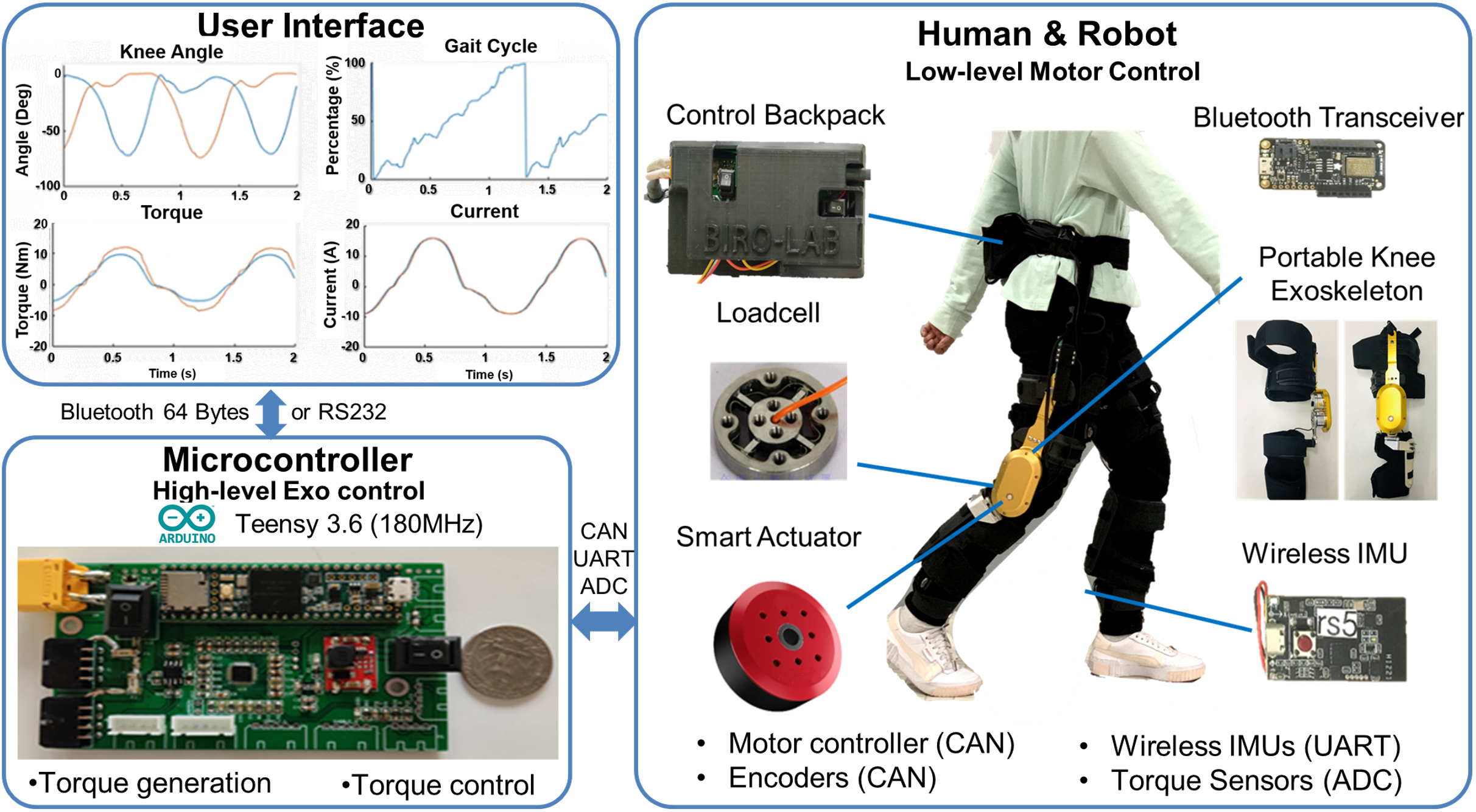}
    \caption{The electronic hardware architecture of the pediatric knee exoskeleton. We implemented high-level control on a Teensy microcontroller. The robot wirelessly communicates with a remote PC for signal monitoring, data collection, and parameter tuning in real-time. IMUs are used to perform gait detection in real-time.}
    \label{fig:Electronics_Hardware_Architecture}
\end{figure}


\subsection{High Torque Density Motor for Compliant Actuator}\label{sec:High_Torque_Density_Motor}
To have a lightweight and high torque actuator for pediatric knee exoskeleton, we customized a brushless DC motor which has $50$ mm diameter and weighs $112$ g. This motor is tailored for pediatric application, while our previous motor design in a hip exoskeleton \cite{yu2020quasi} was for adults, and the motor has $87$ mm diameter and weighs $274$ g, and the actuator weighs $777$ g. The motor can output continuous torque of $0.192$ Nm and weighs $112$ g. The custom motor is connected through a pair of $1:1$ spur gear to a $36:1$ planetary gear assembly, enabling a peak output torque of $20$ Nm and peak output speed of $17.4$ rad/s with a nominal $42$ V supply. The use of spur gear allowed the motor and actuator to be connected in parallel, and thus ensuring the actuator better fit the human body. The actuator assembly weighs $530$ g , and the unilateral exoskeleton weighs $1.65$ kg (including electronics, battery and wearable), giving the complete assembly an exoskeleton torque density of $12.12$ Nm/kg. The transmission ratio in this design is larger than is typically considered a quasi-direct drive design but is small enough to result in low output inertia ($128.3\, \textrm{kg}\cdot \textrm{cm}^2$, see TABLE \ref{tab:Comparison_Pediatric_Exoskeletons}). For comparison, the pediatric exoskeleton in \cite{chen2018design} has output inertia of $779.5$ $\textrm{kg}\cdot \textrm{cm}^2$ with a gear ratio of $153:1$ and weighs $2.59$ kg for a torque density of $5.79$ Nm/kg. Our actuators contain a 14-bit magnetic rotary encoder to measure the rotor position, as well as an embedded microcontroller (STM32F407) which executes low-level motor control. 

\subsection{Electronics and Sensor System}
 We mounted wireless IMU sensors on the thigh and shrank of each leg to measure biological knee angle during walking. To obtain feedback signal for motor control, we used two customized loadcells to measure motor generated torque. These sensors were connected to a custom circuit board which handled high-level torque control for the exoskeleton. The microcontroller (Teensy 3.6, $180$ MHz) executed the outer two loops of a three-stage controller hierarchy. It handled high-level walking control (Section \ref{sec:High_Level_Controller}) and mid-level PID torque control and communicated with low-level current controller and encoder assemblies mounted to each actuator (Section \ref{sec:High_Torque_Density_Motor}) using the CAN bus protocol. Including the batteries, the overall wearable electronics system weighs $370$ g. The electrical system architecture is illustrated in \figurename~\ref{fig:Electronics_Hardware_Architecture}.

\subsection{Control Strategy}\label{sec:High_Level_Controller}
Traditional high-level actuator control methods are usually finite-state machine based \cite{walsh2006autonomous} or time-based \cite{seo2015new}, which rely on accurate gait cycle segmentation and activity classification, or have difficulty in dealing with irregular gait patterns such as those observed in the cerebral palsy population. To provide a robust non-time domain torque profile to assist both flexion and extension, we applied an angle-based control algorithm \cite{lim2019delayed} which can cope with different situations without the need to perform gait phase estimation or activity recognition. The only state variable required consists of the knee angles from both left and right knee, which are calculated as the difference in angle between the corresponding thigh and shank segment. The angle-time curve of the thigh and shank segment from both sides is calculated using the gyroscope readings from four wireless IMUs. With the two knee angles, we define an intermediate state $y_{raw} (t)$ as follows:
\begin{equation}
    y_{raw}(t) = \sin{q_r(t)} - \sin{q_l(t)}
    \label{eqn:Sin_Angle_Difference}
\end{equation}

\noindent where $q_r(t)$ and $q_l(t)$ are measured right and left knee angle. The signal is smoothed by applying a first-order low-pass filter to obtain the smoothed angle difference:
\begin{equation}
    y(t) = (1-\alpha)y(t-1)+\alpha y_{raw}(t),\quad 0<\alpha<1
    \label{eqn:Smoothed_Angle_Difference}
\end{equation}

\noindent where $\alpha$ is a smoothing factor that can be tuned to adjust the relative weight between the previously smoothed angle difference $y(t-1)$ and current measurement $y_{raw}(t)$. Here, we choose $\alpha=0.04$, along with a sampling time $\Delta T=0.001$ sec, which corresponds to a cut-off frequency $f_c$ that is suitable for pediatric walking assistance:
\begin{equation}
    f_c = \frac{\alpha}{(1-\alpha)2\pi\Delta T}\approx 6.63\,\textrm{Hz}
    \label{eqn:Control_Cutoff_Frequency}
\end{equation}

Output torque $\tau(t)$ is obtained by multiplying smoothed angle difference $y(t)$ with an appropriate gain $\kappa$ and applying a time shift $\Delta t$:
\begin{equation}
    \tau(t)=\kappa y(t-\Delta t)
    \label{eqn:Control_Torque_Profile}
\end{equation}

When $\kappa>0$, the output torque assists the subject motion, whereas the output torque is resistive when $\kappa<0$. The time shift $\Delta t$ can be adjusted so that the output torque profile best fits the subject's motion and performed activities. This method works on both legs as the torque applied on the left side is the opposite of that applied on the right side:
\begin{equation}
    \tau_{l,assist}=-\tau_{r,assist}
    \label{eqn:Control_Torque_Left_And_Right}
\end{equation}

\begin{figure}[!htbp]
  \centering
  \begin{subfigure}{\columnwidth}
  \centering
    \includegraphics[width=0.95\columnwidth]{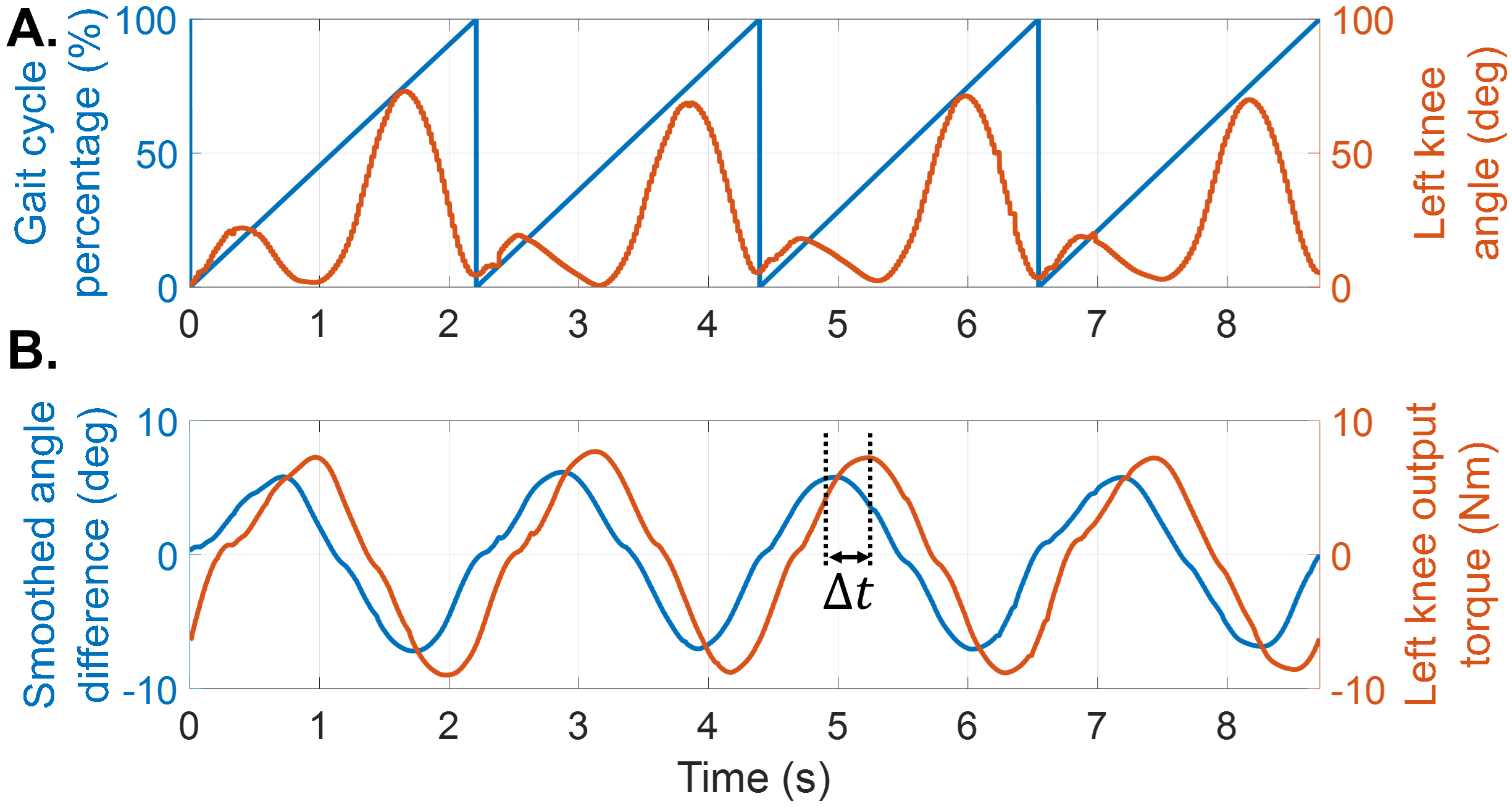}
    \phantomsubcaption
    \label{fig:Control_Gait_Cycle}
  \end{subfigure}
  
  \vspace{-0.4cm}
  \begin{minipage}{\columnwidth}
    \phantomsubcaption
    \label{fig:Control_Torque_Profile}
  \end{minipage}
  
  \caption{(A) Measured left knee angle for a continuous four strides, overlaid with gait cycle percentage. (B) Smoothed angle difference  $y(t)$ resulting from the left knee angle and the right knee angle (not shown here), overlaid with the left knee output torque $\tau_l (t)$ that is time-shifted with respect to $y(t)$.}
  \label{fig:High_Level_Control_Strategy}
\end{figure}

We show an example of walking assistance torque profile generated by $1.0$ m/s treadmill walking in \figurename~\ref{fig:Control_Torque_Profile}, with $\kappa=10$, $\Delta t=0.25$ sec. We also show the corresponding gait cycle and left knee angle for comparison purposes in \figurename~\ref{fig:Control_Gait_Cycle}.

\section{Experiments and Mechatronics Evaluation}
To systematically evaluate the performance of our exoskeleton system, we tested backdrivability, bandwidth, torque tracking, walking, and acoustics. In the backdrivability, torque tracking, and acoustic sound tests, two able-bodied subjects (one 27-year-old, $160$ cm, $50$ kg female: and one 27 year-old, $170$ cm, $60$ kg male) wore our larger size pediatric knee exoskeleton and walked on a treadmill at $1$ m/s. 

\subsection{High Backdrivability}\label{sec:Experiment_Backdrivability}
\begin{figure*}[!htbp]
  \centering
  \begin{subfigure}{\linewidth}
  \centering
    \includegraphics[width=\columnwidth]{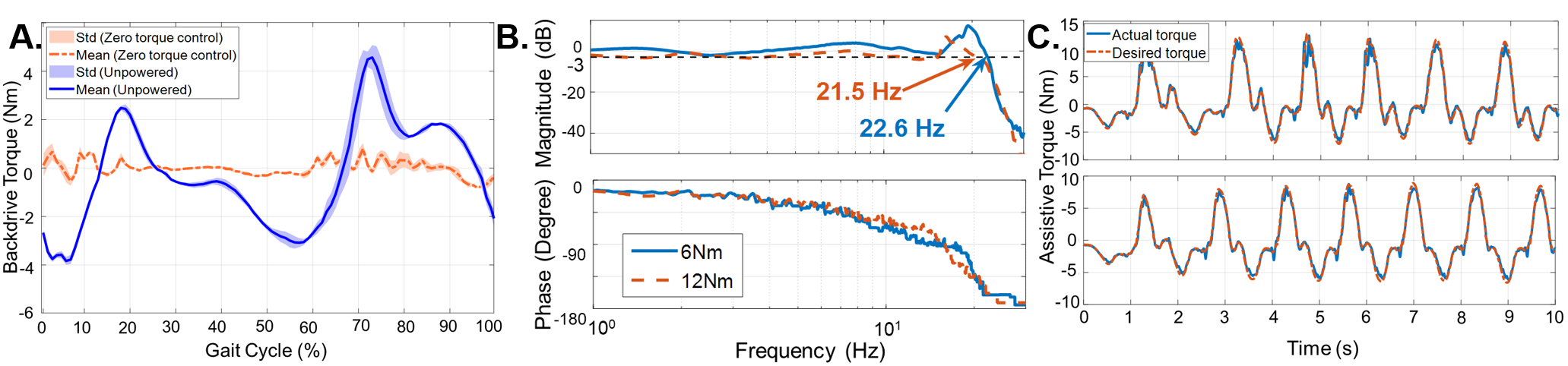}
    \phantomsubcaption
    \label{fig:Backdrive_Torque}
  \end{subfigure}
  
  \vspace{-0.6cm}
  \begin{minipage}{\linewidth}
    \phantomsubcaption
    \label{fig:Backdrive_Bode_Plot}
  \end{minipage}
  
  \vspace{-0.6cm}
  \begin{minipage}{\linewidth}
    \phantomsubcaption
    \label{fig:Backdrive_Torque_Tracking}
  \end{minipage}
  
  \caption{(A) Measured backdrive torque was small in both zero torque control mode (orange) and unpowered mode (blue). (B) Bode plot of closed-loop torque tracking indicates the exoskeleton exhibits high bandwidth with a sinusoidal chirp reference from $1$ to $30$ Hz.  Bandwidth was defined as the frequency at which the torque magnitude fell below $-3$ dB. (C) Torque tracking results in the time domain demonstrate that the actuator is capable of tracking the reference torque for the selected walking speed without noticeable phase lag. Heel-strike causes a minor disturbance of about $1$ Nm in the output torque, which becomes a proportionally less significant source of tracking error as the torque assistance level increases.}
  \label{fig:Backdrive_Test}
  \vspace{-0.5cm}
\end{figure*}

To evaluate the backdrivability of exoskeleton (i.e., reflected inertia and transmission friction) we measured the dynamic backdrive torque under the unpowered and powered zero torque control conditions. Torque data for the backdrivability tests are shown in \figurename~\ref{fig:Backdrive_Torque}, with statistics from $10$ repetitions of the experiment. 

Maximum backdrive torque for the unpowered condition was $4.60 \pm 0.39$ Nm, occurring during peak knee acceleration in the early swing phase. The RMS average torque across the gait cycle was $1.79 \pm 0.03$ Nm. These values primarily represent the reflected inertia of the motor. Maximum backdrive torque for the zero torque control condition was $0.84 \pm 0.14$ Nm, and RMS torque was $0.32$ Nm across the gait cycle. This is a reduction of $82$\% peak torque and $85$\% RMS torque from the unpowered condition. Peak torque for this condition was $7$\% of the expected peak assistance torque of $12$ Nm, indicating a high level of transparency to the user. No clear peaks associated with gait events can be observed in the data for this condition, indicating that the zero-torque controller is successfully compensating for the actuator inertia and friction.

\subsection{High Bandwidth}\label{sec:Experiment_Bandwidth}
Bode plots for the closed-loop torque control frequency response are shown in \figurename~\ref{fig:Backdrive_Bode_Plot}. The bandwidth was $22.6$ Hz and $21.5$ Hz for torque tracking magnitudes of $6$ Nm and $12$ Nm, respectively. Closed-loop torque control bandwidth exceeds that of other pediatric knee exoskeletons \cite{lerner2017lower, chen2018design} (see TABLE \ref{tab:Comparison_Pediatric_Exoskeletons}).

\subsection{Torque Tracking}\label{sec:Experiment_Torque_Tracking}
Closed-loop torque tracking data during 10-second walking trials are illustrated in \figurename~\ref{fig:Backdrive_Torque_Tracking}. RMS tracking errors were $0.46$ Nm ($5.2$\% of peak) and $0.73$ Nm ($5.7$\% of peak) for low and high torque-assist settings, respectively. RMS errors did not exceed $10$\% of peak commanded torque for any condition, indicating high torque tracking accuracy during gait.

\subsection{Average Acoustic Sound Level During Gait}\label{sec:Experiment_Acoustic_Level}
Acoustic sound levels were measured at $57.1$ dB, $68.8$ dB, $72.2$ dB, and $72.4$ dB for the baseline, unpowered, low-torque assistance, and high-torque assistance conditions, respectively. Peak sound levels for all conditions occurred immediately before or after heel strike. The overall difference in sound level between the powered conditions was small, and both are louder than the unpowered condition only by $3.5$ dB on average at their peak volume. In contrast, the increase of $11.7$ dB from the baseline peak to the unpowered peak suggests that the actuators are not the primary contributor to the overall noise level of walking with the exoskeleton. This result agrees with other work that has shown that low-impedance actuators run more quietly than high-impedance actuators performing the same amount of work \cite{nieto2019minimizing}.

\section{User-Specific Actuator Optimization}
We formulate an optimization framework of wearable robot design that incorporates both motor and transmission design for all three kinds of actuation paradigms \cite{yu2020quasi}, namely conventional, series elastic actuation (SEA), and quasi-direct-drive actuation. In particular, our optimization framework is generic for both exoskeleton design and pediatric robot whose size grows during children development.

\begin{figure*}[!hbtp]
  \centering
  \begin{subfigure}{\linewidth}
  \centering
    \includegraphics[width=\linewidth]{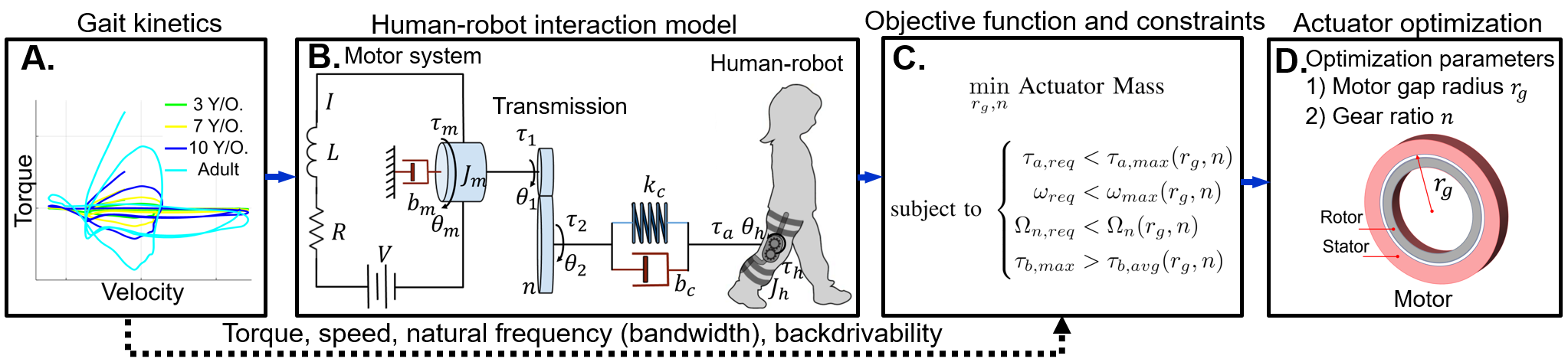}
    \phantomsubcaption
    \label{fig:Optimization_Gait_Kinetics}
  \end{subfigure}
  
  \vspace{-0.5cm}
  \begin{minipage}{\linewidth}
    \phantomsubcaption
    \label{fig:Optimization_HRI_Model}
  \end{minipage}
  
  \vspace{-0.5cm}
  \begin{minipage}{\linewidth}
    \phantomsubcaption
    \label{fig:Optimization_Strategy}
  \end{minipage}
  
  \vspace{-0.5cm}
  \begin{minipage}{\linewidth}
    \phantomsubcaption
    \label{fig:Optimization_Actuator}
  \end{minipage}
  
  \caption{Overview of actuator optimization framework. (A) Pediatric gait kinetics. Our control algorithm prescribes customizable exoskeleton profiles from pathology characterization data. (B) Model of the coupled human-knee exoskeleton system. It included a quasi-direct-drive actuator, wearable structure, and human limbs. (C) Optimization objective function and constraints. We minimized the actuator mechanical mass subject to four constraints, i.e. assistive torque $\tau_{a,req}$, bandwidth $\Omega_{n,req}$, angular velocity $\omega_{n,req}$, backdrive torque $\tau_{b,max}$. (D) Optimization parameters. Our optimization objective function has two optimization parameters, motor gap radius $r_g$ and motor gear ratio $n$. It indicates our method is able to optimize not only the motor but also the transmission.}
  \label{fig:Optimization_Framework}
  \vspace{-0.4cm}
\end{figure*}

\subsection{Actuator Design Optimization Framework}\label{sec:Actuator_Optimization_Framework}
Children of different ages have different requirements in terms of assistive torque, backdrivability, bandwidth and angular velocity from the exoskeleton system. To address these requirements, we identified two key parameters (motor radius and gear ratio) and optimized the actuator mass while treating the above requirements as constraints. While in this work we minimized actuator mass as an case study, other objective functions (e.g., energy consumption) are also feasible using this optimization framework.  \figurename~\ref{fig:Optimization_Framework} shows the overview of our actuator optimization framework that is guided by the torque-velocity profile of children of various ages and takes into consideration of the human-robot interaction model.

The actuator optimization objective function is defined in equation \eqref{eqn:Optimization_Objective_Function} and the constraints are listed in equation \eqref{eqn:Optimization_Constraints}, where $\tau_{a,req}$ is the required maximum torque, $\omega_{req}$ the required maximum angular velocity, $\Omega_{n,req}$ the required natural frequency under torque control, $\tau_{b,max}$ the maximum backdrive torque, $r_g$ the gap radius, and $n$ the gear ratio. 
For equation \eqref{eqn:Optimization_Objective_Function}, we assumed gear ratio has no effect on the gear weight and the motor weight is proportional to the square gap radius $r_g^2$.
\begin{equation}
    \min_{r_g, n}\, \textrm{Actuator Mass} 
    \label{eqn:Optimization_Objective_Function}
\end{equation}
\vspace{-0.2cm}
    \begin{empheq}[left=\text{subject to }\empheqlbrace]{align}
      \tau_{a,req}&< \tau_{a,max}(r_g,n) \nonumber\\
      \omega_{req}&< \omega_{max}(r_g,n) \nonumber\\
      \Omega_{n,req}&< \Omega_n(r_g,n) \nonumber\\
      \tau_{b,max}&> \tau_{b,avg}(r_g,n) \label{eqn:Optimization_Constraints}
    \end{empheq}

\begin{figure}[b]
    \centering
    \includegraphics[width=0.9\columnwidth]{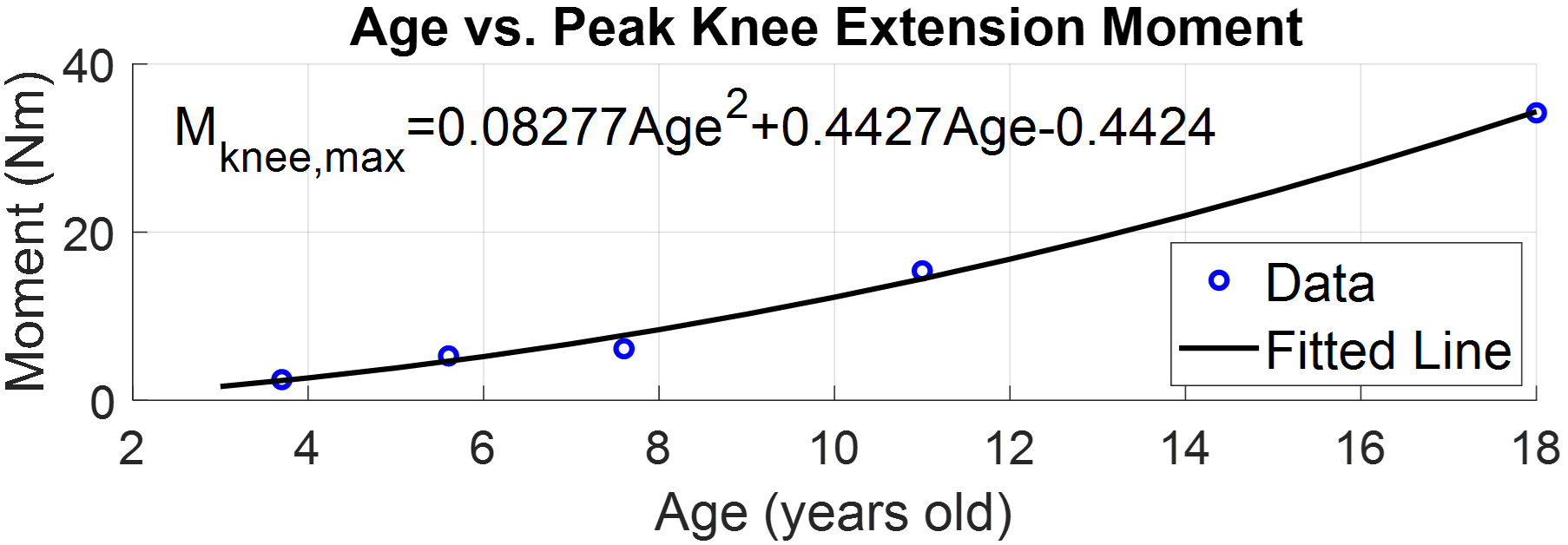}
    \caption{The peak knee extension moment in different ages. The blue dots are the peak knee moments from  \cite{chester2006comparison} (the  dataset for 3-13 years old children) and \cite{fukuchi2018public} (the  dataset  for  adults). The black line is the fitted quadratic function.}
    \label{fig:Peak_Knee_Moment_vs_Age_Fitting}
\end{figure}

We further assumed the required torque $\tau_{a,req}$ was age-specific and related to the peak knee extension moment. We defined peak knee extension moment using publicly available datasets of overground walking measured at self-selected walking speeds for $3-13$ years old children \cite{chester2006comparison} and  adults \cite{fukuchi2018public}. As shown in \figurename~\ref{fig:Peak_Knee_Moment_vs_Age_Fitting}, a quadratic curve was used to fit the relationship between the age $A$ and peak knee extension moment $M_{knee,max}$ where the age for the adult dataset was set as $18$. The fitted curve is shown in equation \eqref{eqn:Optimization_Result_Max_Knee_Torque}. The required torque $\tau_{a,req}$ was set at 30\% of the peak knee extension moment with a safety factor of $2$, shown as equation \eqref{eqn:Optimization_Result_Required_Torque}.
\begin{align}
    M_{knee,max}&=0.08277\textrm{Age}^2+0.4427\textrm{Age}-0.4424\label{eqn:Optimization_Result_Max_Knee_Torque}\\
    \tau_{a,req}&=0.3M_{knee,max}\times2\label{eqn:Optimization_Result_Required_Torque}
\end{align}

\vspace{-0.5cm}
The required angular velocity $\omega_{req}$ was set as $2\pi$ rad/s which corresponds to the maximum angular velocity under $1$-Hz walking cycle from the datasets \cite{chester2006comparison, fukuchi2018public}. The required natural frequency $\Omega_{n,req}$ was set as $20$ Hz for human walking. The maximum backdrive torque $\tau_{b,max}$ of the unpowered exoskeleton in $1$-Hz walking cycle was set as $5$ Nm. 

\subsection{Human-Robot Interaction Model of Torque Control}\label{sec:Optimization_Human_Exoskeleton_Model}
\begin{figure}[!htbp]
  \centering
  \begin{subfigure}{\columnwidth}
  \centering
    \includegraphics[width=0.95\columnwidth]{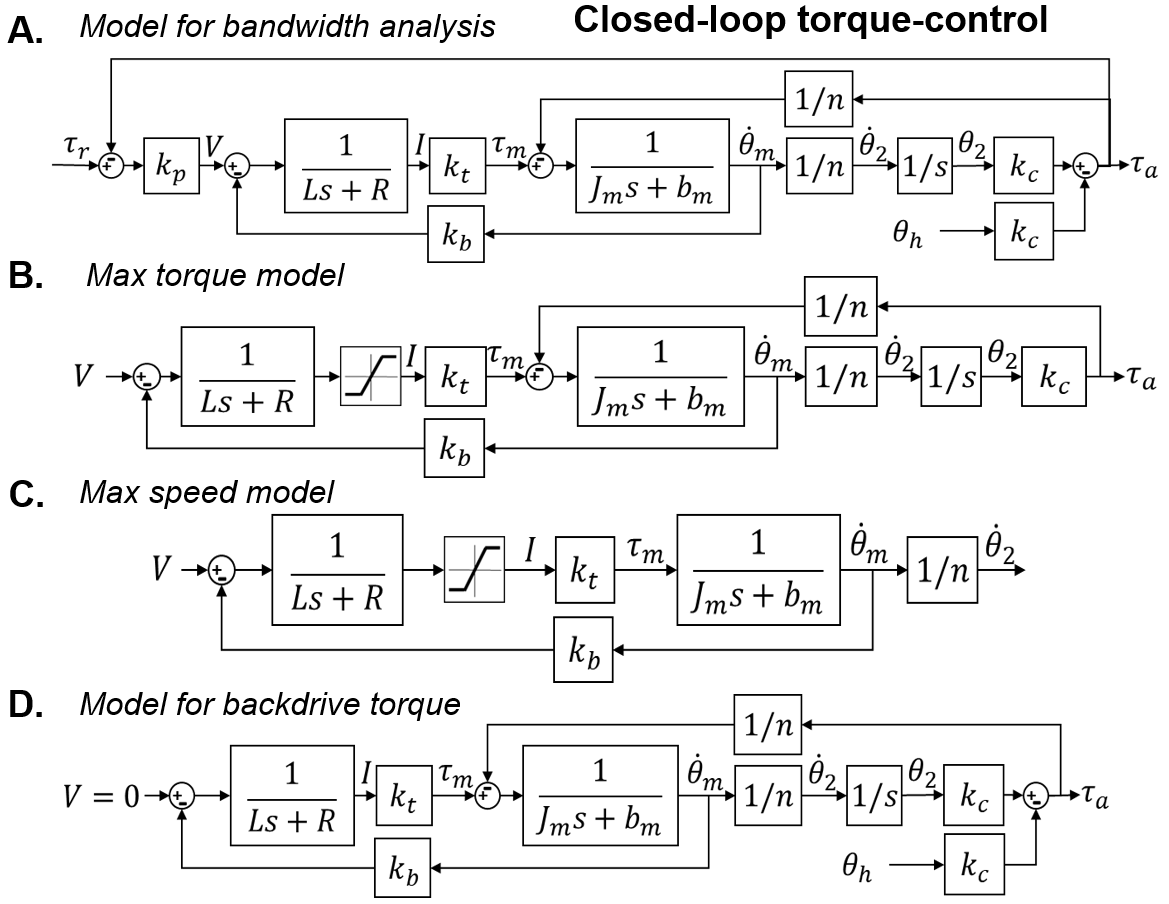}
    \phantomsubcaption
    \label{fig:Exo_Block_Closed_Loop}
  \end{subfigure}
 
  \vspace{-0.5cm}
  \begin{minipage}{\columnwidth}
    \phantomsubcaption
    \label{fig:Exo_Block_Max_Torque}
  \end{minipage}
  
  \vspace{-0.5cm}
  \begin{minipage}{\columnwidth}
    \phantomsubcaption
    \label{fig:Exo_Block_Max_Speed}
  \end{minipage}
  
  \begin{minipage}{\columnwidth}
    \phantomsubcaption
    \label{fig:Exo_Block_Backdrive}
  \end{minipage}
  
  \caption{Block diagrams for the exoskeleton dynamic system in various model configurations: Model for (A) closed-loop torque control, (B) maximum torque, (C) maximum speed, and (D) backdrive torque and bandwidth.}
  \label{fig:Exo_Block_Diagrams}
\end{figure}

Our proposed optimization framework took into consideration of the human-robot interaction. The model of the human-exoskeleton system and block diagram of torque control were modeled as \figurename~\ref{fig:Optimization_HRI_Model} and \figurename~\ref{fig:Exo_Block_Closed_Loop}. The human exoskeleton system incorporated the electromechanical model of the quasi-direct-drive actuator, wearable structure, and human limbs. The nomenclature included $V$ winding voltage, $L$ terminal inductance of motor, $R$ terminal resistance of the motor, $I$ motor current, $J_m$ moment of inertia of the rotor, $\theta_m$ motor angle, $\tau_m$ motor output torque, $b_m$ motor damping coefficient, $\tau_1$ gear input torque, $\theta_1$ gear input angle, $n$ gear ratio, $\tau_2$ gear output torque, $\theta_2$ gear output angle, $k_c$ human-exoskeleton transmission stiffness, $b_c$ human-exoskeleton transmission damping, $\tau_a$ output assistive torque, $\tau_h$ knee total muscle torque, $J_h$  moment of inertia of human shank, $\theta_h$ knee angle, $k_p$ the $P$ gain, and $k_i$ the $I$ gain. 

The block diagram of the torque control shown as \figurename~\ref{fig:Exo_Block_Closed_Loop} was implemented to assist human walking according to the knee joint angles (Section \ref{sec:High_Level_Controller}). The input was the torque reference $\tau_r$ and the output was the actual torque $\tau_a$ applied to human. The transmission damper coefficient $b_c$ was small and set as zero. To investigate the bandwidth of the torque control, the input human knee angle $\theta_h$ was set to zero. The closed-loop transfer function of torque control was derived as equation \eqref{eqn:Closed_Loop_Torque_Control}, where $I$ gain $k_i$ was set to zero for simplicity, and winding inductance $L$ was set to $0$ due to its small value.


\begin{equation}
    \scalebox{1.1}{$\left.\frac{\tau_a(s)}{\tau_r(s)}\right\vert_{\substack{\theta_h(s)=0,\\ L=0}} = \frac{k_p k_c k_t n}{n^2 R J_m s^2 + n^2(R b_m+k_b k_t)s+k_c(R+k_p k_t n)}$}
    \label{eqn:Closed_Loop_Torque_Control}
\end{equation}


The natural frequency $\Omega_n$ of closed-loop control is analytically derived in equation \eqref{eqn:Natural_Frequency_Torque_Control}. It increases with decreasing gear ratio, moment of inertia, or increasing torque constant.
\begin{equation}
    \Omega_{n, torque\,control}=\sqrt{\frac{k_c(R+k_p k_t n)}{n^2 R J_m}}
    \label{eqn:Natural_Frequency_Torque_Control}
\end{equation}

The backdrivability was estimated by the backdrive torque under the unpowered condition. The block diagram is shown in \figurename~\ref{fig:Exo_Block_Backdrive}. The input is the human movement $\theta_h$ and the output is the backdrive torque $\tau_a$ with the input voltage supply $V$ equal to $0$. As the inductance is small, the transfer function reduces to equation \eqref{eqn:Backdrive_Transfer_Function}.
\begin{equation}
    \left.\frac{\tau_a(s)}{\theta_h(s)}\right\vert_{V=0} = \frac{-k_c n^2 s[J_m Rs + (R b_m+k_b k_t)]}{n^2 s[J_m R s+(R b_m+k_b k_t)]+R k_c}
    \label{eqn:Backdrive_Transfer_Function}
\end{equation}

As human motion is mostly low frequency ($\omega\rightarrow0$) and the gear ratio $n$ is small, the magnitude of  $\left|\frac{\tau_a(s)}{\theta_h(s)}\right|_{s=j\omega}$ goes to zero with decreasing $\omega$ and $n$. Therefore the system with lower frequency, smaller motor, and lower gear ratio produces lower backdrive torque and better backdrivability.
\begin{table}
  \centering
  \caption{The Relationship Between Motor Parameters and Gap Radius} \label{tab:Motor_Parameters_Gap_Radius}
  \begin{tabular}{C{1.0cm} C{3.2cm} C{2.1cm} C{1.1cm}}\toprule
    Symbols & Motor Parameters & Relationship with Gap Radius ($r_g$) & Our Motor \\\hline
    $r_g$ & Gap radius (m) & - & 0.021 \\
    $r_m$ & Motor radius (m) & $r_m\propto r_g$ & 0.026 \\
    $M_m$ & Mass (kg) & $M_m\propto r_g^2$ & 0.112 \\
    $J_m$ & Moment of inertia (kg$\cdot \textrm{m}^2$) & $J_m\propto r_g^3$ & $9.9\textrm{E-6}$ \\
    $b_m$ & Motor damping (Nm$\cdot$s/rad) & - & 0.01 \\
    $k_t$ & Torque constant (Nm/A) & $k_t\propto r_g$ & 0.04 \\
    $k_b$ & Back EMF constant (V$\cdot$s/rad) & $k_b\propto r_g$ & 0.04 \\
    $R$ & Terminal resistance ($\Omega$) & $R\propto r_g^{-1}$ & 0.74 \\
    $L$ & Terminal inductance (H) & $L\propto r_g^{-1}$ & $2.98\textrm{E-4}$ \\
    $V_m$ & Maximum voltage (V) & - & 42 \\
    $I_{max}$ & Maximum current (A) & $I_{max}\propto r_g$ & 16.5 \\
    $\tau_m$ & Motor torque (Nm) & $\tau_m\propto r_g^2$ & - \\
    $\tau_{m,max}$ & Maximum motor torque (Nm) & $\tau_{m,max}\propto r_g^2$ & 0.66 \\\hline
  \end{tabular}
\end{table}

In summary, a system with low backdrive torque and high natural frequency is the most suitable for an exoskeleton. However, these design variables all have trade-offs concerning the performance parameters. Maximizing backdrivability requires a low gear ratio and small motor with low moment of inertia and damping coefficient. Maximizing output torque requires a high gear ratio and high torque constant. Maximizing speed requires a low gear ratio and low back-EMF constant. 
To form the basis of this optimization, we quantify the effect of motor size and gear ratio on the output torque, speed, bandwidth, and backdrive torque in the following section.

\subsection{Optimization Parameters for Motor and Transmission}
\subsubsection{Geometry Consideration of High Torque Density Motors}
According to \cite{wensing2017proprioceptive}, the performance of a high torque density motor with a small moment of inertia can be described by the gap radius of the motor with fixed rotor and stator radial thickness. The relationship between gap radius and the motor parameters is shown in TABLE \ref{tab:Motor_Parameters_Gap_Radius}, with the parameters of our prototype motor provided as a point of reference. 

\subsubsection{Gear Ratio Consideration of the Exoskeleton System}
From the conclusion of section \ref{sec:Optimization_Human_Exoskeleton_Model}, the relationship between the motor performance and the gear ratio is shown in TABLE \ref{tab:Motor_Parameters_Gear_Ratio}. As the gear ratio increases, the output torque increases, while the output speed, bandwidth of torque control, backdrivability decreases.
\begin{table}
  \centering
  \caption{The Relationship Between Motor Performance and Gear Ratio} \label{tab:Motor_Parameters_Gear_Ratio}
  \begin{tabular}{ccc}\toprule
    Symbols & Motor Parameters & Relationship with Gear Ratio ($n$) \\\hline
    $\tau_{a,max}$ & Maximum output torque & $\tau_{a,max}\propto n$ \\
    $\dot{\theta}_{2,max}$ & Maximum output speed & $\dot{\theta}_{2,max}\propto 1/n$\\
    - & Bandwidth & $\propto 1/n$\\
    - & Maximum backdrive torque & $\propto 1/n$\\\hline
  \end{tabular}
\end{table}

\subsection{Optimization Constraints}\label{sec:Optimization_Constraints}
The motor performance for the actuation optimization are analyzed in the following five aspects: 1) the maximum output torque, 2) the maximum output speed, 3) bandwidth of torque control, 4) backdrive torque, and 5) mechanism weight. There are trade-off between motor performance and motor gap radius and gear ratio, which is illustrated in the following section.


\subsubsection{Constraint of Maximum Output Torque}
\begin{figure}[!htbp]
  \centering
  \begin{subfigure}{\columnwidth}
  \centering
    \includegraphics[width=0.9\columnwidth]{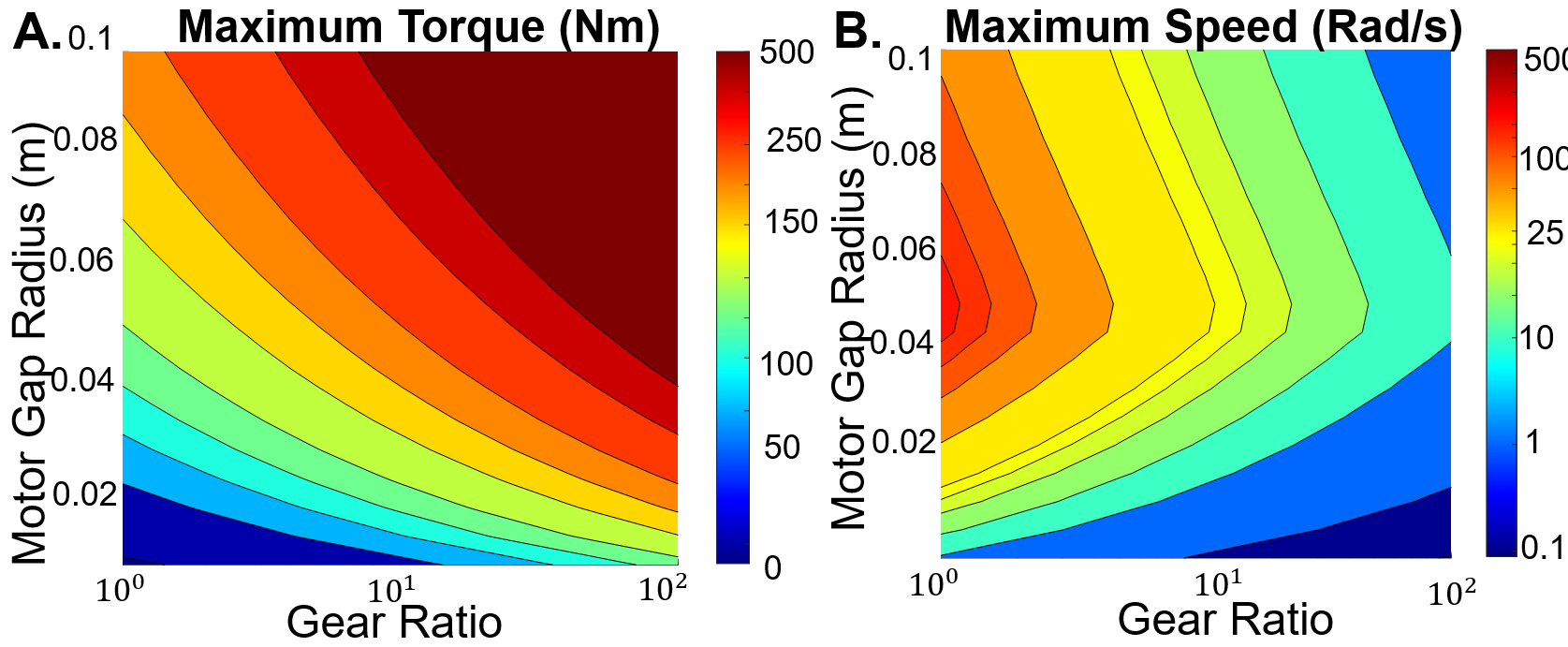}
    \phantomsubcaption
    \label{fig:Gap_Radius_vs_Gear_Ratio_Max_Torque}
  \end{subfigure}
  
  \vspace{-0.3cm}
  \begin{minipage}{\columnwidth}
    \phantomsubcaption
    \label{fig:Gap_Radius_vs_Gear_Ratio_Max_Speed}
  \end{minipage}
  
  \caption{(A) The result of the maximum output torque. It shows that the maximum torque increases as the gear ratio and motor gap radius increases. (B) The result of maximum speed in different gap radius and gear ratio. With the same gear ratio, the maximum angular velocity is obtained as the gap radius nears $0.05$ meter. As the gear ratio increases, the maximum angular velocity decreases.}
  \label{fig:Gap_Radius_vs_Gear_Ratio}
  \vspace{-0.2cm}
\end{figure}

To estimate the maximum output torque related to the different gap radius and gear ratio, the block diagram was derived in \figurename~\ref{fig:Exo_Block_Max_Torque} assuming fixed output angle. 
The input voltage $V$ was set to maximum voltage $V_{max}=42\,V$ and the peak value of the output torque $\tau_a$ was the maximum output torque $\tau_{a,max}$. The saturation current was set to maximum current $I_{max}$ corresponding to the different gap radius. The result is shown in \figurename~\ref{fig:Gap_Radius_vs_Gear_Ratio_Max_Torque}. When the gear ratio and the gap radius increases, the maximum torque increases. The $\tau_{a,max} (r_g,n)$ in equation \eqref{eqn:Optimization_Constraints} is obtained by finding the maximum output torque in the figure given a specific gap radius and gear ratio.

\vspace{-0.1cm}

\subsubsection{Constraint of Maximum Output Speed}
To estimate the maximum output velocity related to the different gap radius and gear ratio, the block diagram was derived in \figurename~\ref{fig:Exo_Block_Max_Speed} assuming free output rotation. The input voltage $V$ was set as $V_{max}=42\,V$ and the peak output angular velocity $\omega$ was the maximum output angular velocity $\omega_{max}$. The result is shown in \figurename~\ref{fig:Gap_Radius_vs_Gear_Ratio_Max_Speed}. When the gear ratio  increases, the maximum angular velocity decreases. The maximum angular velocity is obtained as the gap radius nears $0.05$ meters assuming constant gear ratio. The $\omega_{max} (r_g,n)$ in equation \eqref{eqn:Optimization_Constraints} is obtained by finding the maximum output speed in the figure given a specific gap radius and gear ratio.

\subsubsection{Constraint of Natural Frequency of Torque Control}
\begin{figure}[!htbp]
  \centering
  \begin{subfigure}{\columnwidth}
  \centering
    \includegraphics[width=0.9\columnwidth]{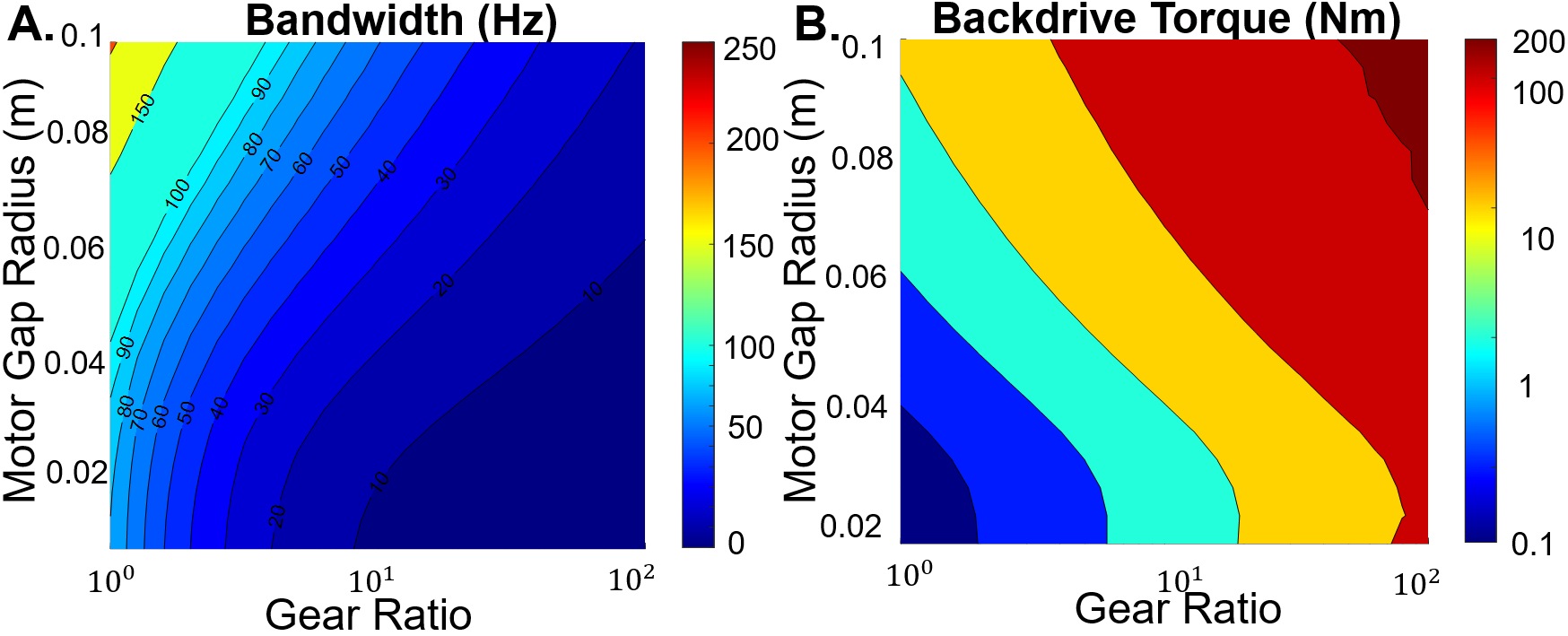}
    \phantomsubcaption
    \label{fig:Gap_Radius_vs_Gear_Ratio_Bandwidth}
  \end{subfigure}
  
  \vspace{-0.3cm}
  \begin{minipage}{\columnwidth}
    \phantomsubcaption
    \label{fig:Gap_Radius_vs_Gear_Ratio_Backdrive}
  \end{minipage}
  
  \caption{Torque control bandwidth and backdrive torque with different motor gap diameter and gear ratio (A) High frequency torque control requires a low gear ratio and high motor gap diameter. (B) Low resistance torque (high backdrivability) is ensured by a small motor gap diameter and low gear ratio.}
  \label{fig:Gap_Radius_vs_Gear_Ratio_2}
  \vspace{-0.00cm}
\end{figure}

The bandwidth of torque control with different gap radius and gear ratio depends on the natural frequency. The natural frequency $\Omega_n$ as appeared in equation \eqref{eqn:Natural_Frequency_Torque_Control} is shown in \figurename~\ref{fig:Gap_Radius_vs_Gear_Ratio_Bandwidth}, where $P$ gain $k_p$ is $1$. When the gap radius decreases and gear ratio increases, natural frequency of torque control decreases. The $\Omega_{n} (r_g,n)$ in equation \eqref{eqn:Optimization_Constraints} is obtained by finding the bandwidth corresponding to a $1$-Hz walking cycle in the figure given a specific gap radius and gear ratio.

\subsubsection{Constraint of Backdrivability}
The average backdrive torque $\tau_{b,avg}$ is simulated by the block diagram, shown in \figurename~\ref{fig:Exo_Block_Backdrive} and the result is shown in \figurename~\ref{fig:Gap_Radius_vs_Gear_Ratio_Backdrive}. The input is the trajectory of the knee angle $\theta_h$ in the $1$-Hz walking cycle. When the gap radius and the gear ratio increases, the backdrive torque increases. The $\tau_{b,avg} (r_g,n)$ in equation \eqref{eqn:Optimization_Constraints} is obtained by finding the backdrive torque corresponding to a $1$-Hz walking cycle in the figure given a specific gap radius and gear ratio.


\subsection{Constrained Optimization Results}
\begin{figure}[!htbp]
  \centering
  \begin{subfigure}{\columnwidth}
  \centering
    \includegraphics[width=0.95\columnwidth]{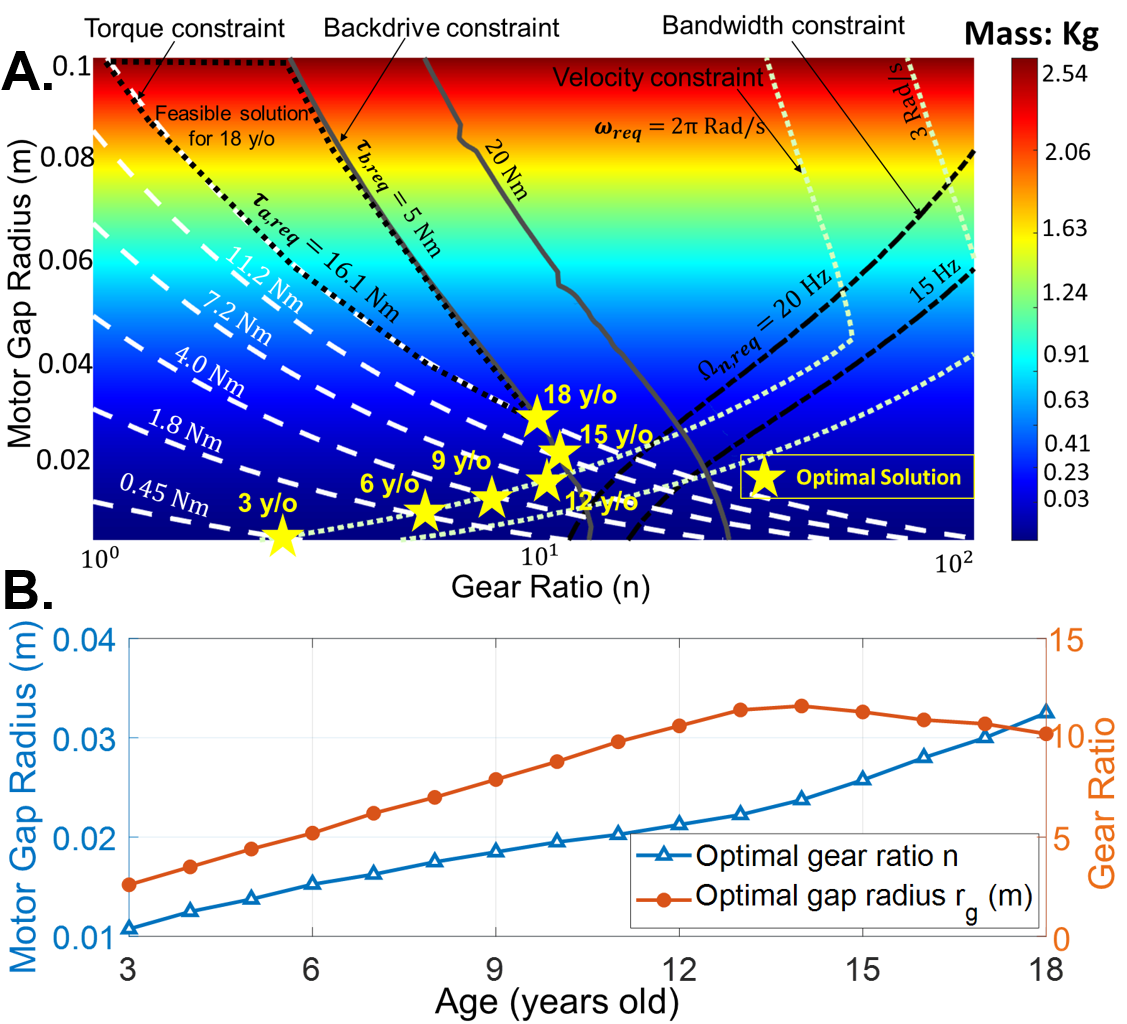}
    \phantomsubcaption
    \label{fig:Optimization_Result_Contour}
  \end{subfigure}
  
  \vspace{-0.3cm}
  \begin{minipage}{\columnwidth}
    \phantomsubcaption
    \label{fig:Optimization_Result_Gap_Radius_vs_Age}
  \end{minipage}
  
  \caption{(A) Constraint contours in the optimization parameter space. The age-specific optimal solutions are found at the point that has the smallest mass in the region bounded by the four constraint contours. (B) The optimal gap radius increases monotonically for children aging from $3$ to $18$, while optimal gear ratio increases for age between 3 and 13 years old and then decreases for age between 13 and 18 years old.}
  \label{fig:Optimization_Result}
\end{figure}

The objective function and the constraint contours were plotted in \figurename~\ref{fig:Optimization_Result_Contour}. The minimum value of the objective function was found in the area bounded by the four constraint contours. For example, the optimal solution for $18$ years old was found at the intersection point of the required torque contour ($\tau_{a,req}=16.1$ Nm)  and the maximum backdrive torque contour ($\tau_{b,max}=5$ Nm). This also led to minimal actuator mass as it corresponds to the smallest gap radius.

Finally, the age-related optimal result is shown in \figurename~\ref{fig:Optimization_Result_Gap_Radius_vs_Age}. It showed the optimal motor gap radius monotonically increased from $0.011$ to $0.033$ m when the age increased. The optimal gear ratio increased from $2.6$ to $11.6$ as the age increased from $3$ to $13$ years old. When the age is between $13$ and $18$, the optimal gear ratio decreased from $11.6$ to $10.2$ while the optimal gap radius still increases. It is because the backdrive torque increased dramatically as the gear ratio increases (see \figurename~\ref{fig:Optimization_Result_Gap_Radius_vs_Age}). In summary, we demonstrated the principle to find the optimal exoskeleton design parameters for children of different ages by minimizing the actuator weight and satisfying the requirements for the maximum torque, maximum speed, natural frequency, and backdrive torque.

\section{Discussion and Conclusion}
We developed a bilateral exoskeleton with low mass, high compliance, high bandwidth, and high torque density that can potentially aid with crouch gait mobility enhancement for children with cerebral palsy. Our proposed actuator optimization framework, which considered both motor and transmission design, could be applied to multiple control paradigms and various wearable robots. We were able to meet our peak torque and bandwidth design requirements as determined by biomechanical analysis of cerebral palsy gait pathology and pediatric rehabilitation needs. However, the scope of the presented work is limited to testing only on able-bodied subjects, and the efficacy of the exoskeleton on correcting crouch gait in children is yet untested. Furthermore, by using a high torque-density motor we had previously designed, we needed a $36:1$ gear ratio to achieve the required torque, which is higher than typically preferred for quasi-direct drive actuator designs. However, this gear ratio is still relatively low compared to existing rehabilitation exoskeletons, and as a result, we substantially reduced the energy losses, reflected inertia, and actuator noise associated with large gear reductions typically required for wearable robots using conventional motors. The resulting device has the best performance among state-of-the-art single-joint exoskeletons (TABLE \ref{tab:Comparison_Pediatric_Exoskeletons}). Exoskeletons should have a overall weight of less than $2.5$ kg to not alter the lower limb kinematics of children less than $5$ years old \cite{rossi2013feasibility}, and thus any exoskeletons that are suitable for children younger than $5$ years old should be lighter. Furthermore, the smallest gear ratio and actuator inertia means the exoskeleton has the smallest backdrive torque which ensures high compliance.

We also showed that it is possible to further increase device performance if the gear reduction is reduced into the realm of quasi-direct drive actuation. Using our optimization framework, we demonstrated that minimal actuator mass can be achieved while still meeting the required performance of the exoskeleton for gear ratios of $11.6:1$ and lower. The backdrivability tests showed that the original actuator was capable of being highly transparent to the user with torque control enabled, however, it reaches peak backdrive torques of nearly $5$ Nm when unpowered. We hypothesize that the optimized version of this actuator should demonstrate high transparency in both powered and unpowered conditions. Furthermore, we demonstrated that optimal actuator design varies by age, indicating that age-specific actuator designs may be beneficial as user graduates from one motor size to the next. While we used actuator mass as the objective function in the presented work, our optimization framework can also be used to optimize other quantities (e.g., energy consumption).
\begin{table*}[!htbp]
\centering
\begin{threeparttable}
  \caption{Comparison of Pediatric Exoskeletons} \label{tab:Comparison_Pediatric_Exoskeletons}
  \begin{tabular}{C{2.5cm}C{2.2cm}C{2.2cm}C{2.8cm}C{1.5cm}C{2.0cm}C{1.7cm}}
  \toprule
    Pediatric Exoskeleton \newline & Unilateral Weight \newline (kg) & Actuator Torque (Nm) & Exoskeleton Torque Density (Nm/kg) & Gear Ratio \newline & Actuator Inertia$^a$ ($\textrm{kg}\cdot\textrm{cm}^2$) & Bandwidth (Hz) \\\hline
    MIT 2015\cite{michmizos2015robot} & $>10$ & 7.21 & tethered system & 138:1 & 105.50 & medium \\
    NIH 2017\cite{lerner2017lower} & 1.75$^b$ & 16.10 & tethered system & 311.5:1 & 537.56 & medium \\
    NIH 2018\cite{chen2018design} & 2.59 & 15.00 & 5.79 & 153:1 & 779.5 & medium (9) \\
    NAU 2018\cite{lerner2018untethered} & 1.85 & 24.00 & 12.97 & 331:1 & 976.2 & medium \\
    \textbf{Ours} & \textbf{1.65} & \textbf{20.00} & \textbf{12.12} & \textbf{36:1} & \textbf{128.30} & \textbf{high (22.6)} \\\hline
  \end{tabular}
  \begin{tablenotes}
  \item ${^a}\textrm{Actuator inertia} = \textrm{Motor inertia} \times \textrm{gear ratio}^2. \,{^b}$It is the weight of device brace and actuator. \cite{michmizos2015robot} and \cite{lerner2018untethered} are pediatric ankle exoskeletons.
  \end{tablenotes}
\end{threeparttable}
\vspace{-0.4cm}
\end{table*}

Our model of the mechanical and control system allows for the potential addition of controller optimization in tandem with hardware optimization. While our case study focuses on one specific application in pediatric couch gait rehabilitation, our optimization methods could be broadly applicable within wearable robotics.

\section*{Acknowledgment}

The authors would like to thank Dr. Gray C. Thomas and Prof. Elliott J. Rouse from the University of Michigan for helping to edit this manuscript.

\ifCLASSOPTIONcaptionsoff
  \newpage
\fi



%


\bibliographystyle{IEEEtran}
\bibliography{tmech_pediatric}
%

\begin{IEEEbiography}[{\includegraphics[width=1in,height=1.25in,clip,keepaspectratio]{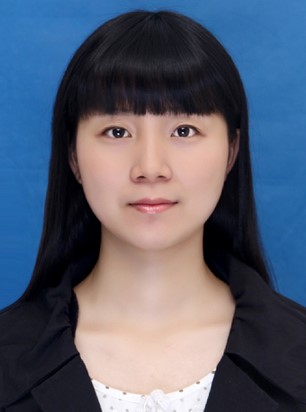}}]{Sainan Zhang}
received B.S. degree in automatic from Xi'an University of Posts and Telecommunications, in 2015, the M.S. degree in control science of engineering from the University of Electronic Science and Technology of China, Sichuan, China, in 2018. She is currently working towards the Ph.D. degree supervised by Dr. Hao Su at the Department of Mechanical Engineering at the City University of New York, New York, USA. Her current research interests include the control and optimization of lower-limb wearable robots.
\end{IEEEbiography}

\begin{IEEEbiography}[{\includegraphics[width=1in,height=1.25in,clip,keepaspectratio]{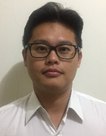}}]{Tzu-Hao Huang}
received the B.S. degree in occupational therapy from National Cheng Kung University, Tainan, Taiwan, in 2004, the M.S. degree in the Institute of Rehabilitation Science and Technology from National Yang Ming University, Taipei, Taiwan, in 2006, and the Ph.D. degree in mechanical engineering from National Taiwan University, Taipei, Taiwan, in 2013.  He is currently an assistant professor in the Department of Mechanical Engineering at the City College of New York. His major research area is control and design of the rehabilitation and assistive devices, human machine interface for assistive and rehabilitation device, brain-machine interface for subjects with movement disability, and the smart textile for physiological sensing in firefight, sport, and medical application.  
\end{IEEEbiography}

\begin{IEEEbiography}[{\includegraphics[width=1in,height=1.25in,clip,keepaspectratio]{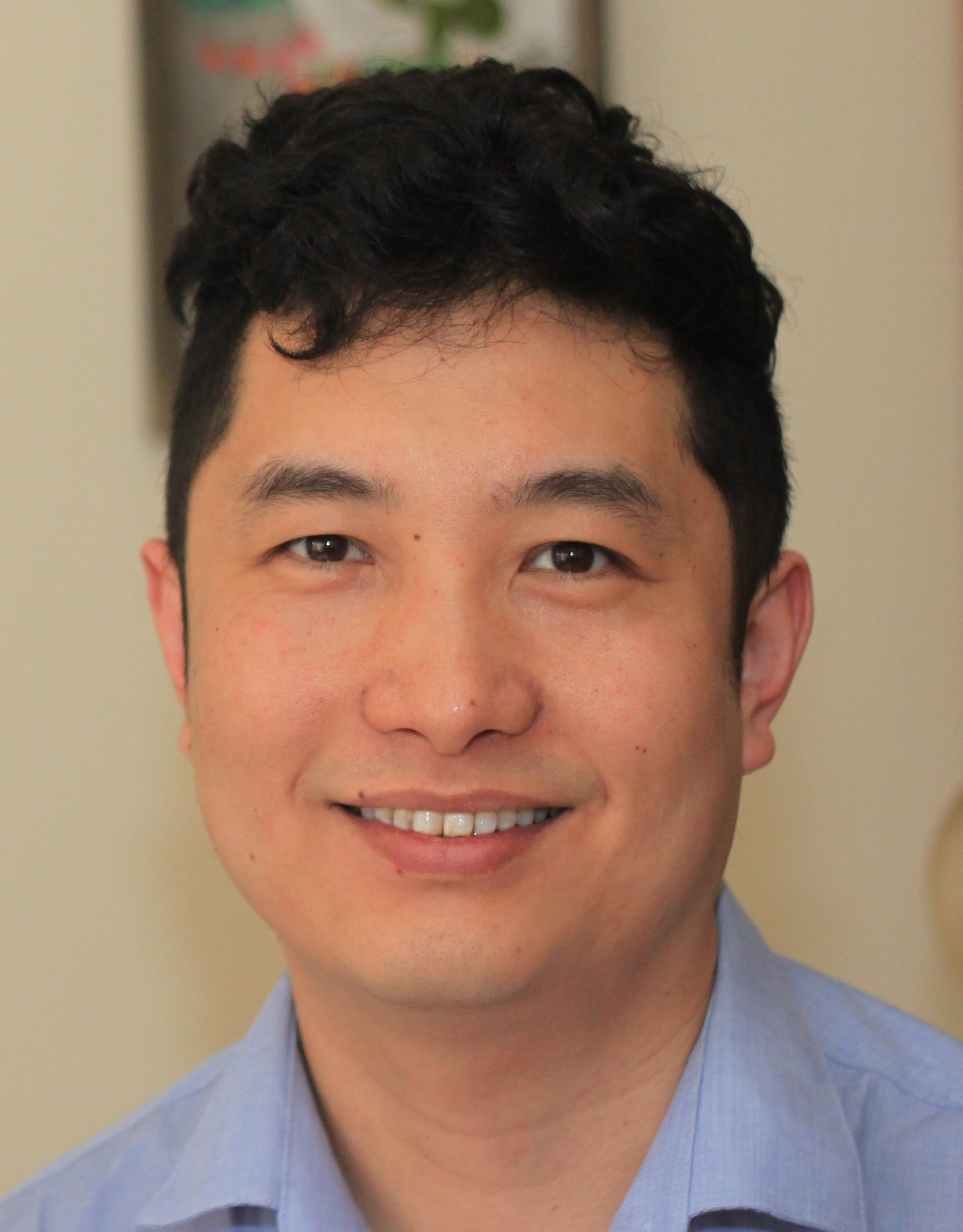}}]{Chunhai Jiao}
received the B.S. degree from Northwest A\&F University, China, and the M.S. degree from the City College of New York, New York, USA. He has been committed to the mechanical design for 16 years in the field of the machine tool, planetary gear transmission and Automation equipment. His current research interests include mechanical design of the lower-limb, upper limb exoskeleton and other experimental devices to support bio robotic projects.
\end{IEEEbiography}

\begin{IEEEbiography}[{\includegraphics[width=1in,height=1.25in,clip,keepaspectratio]{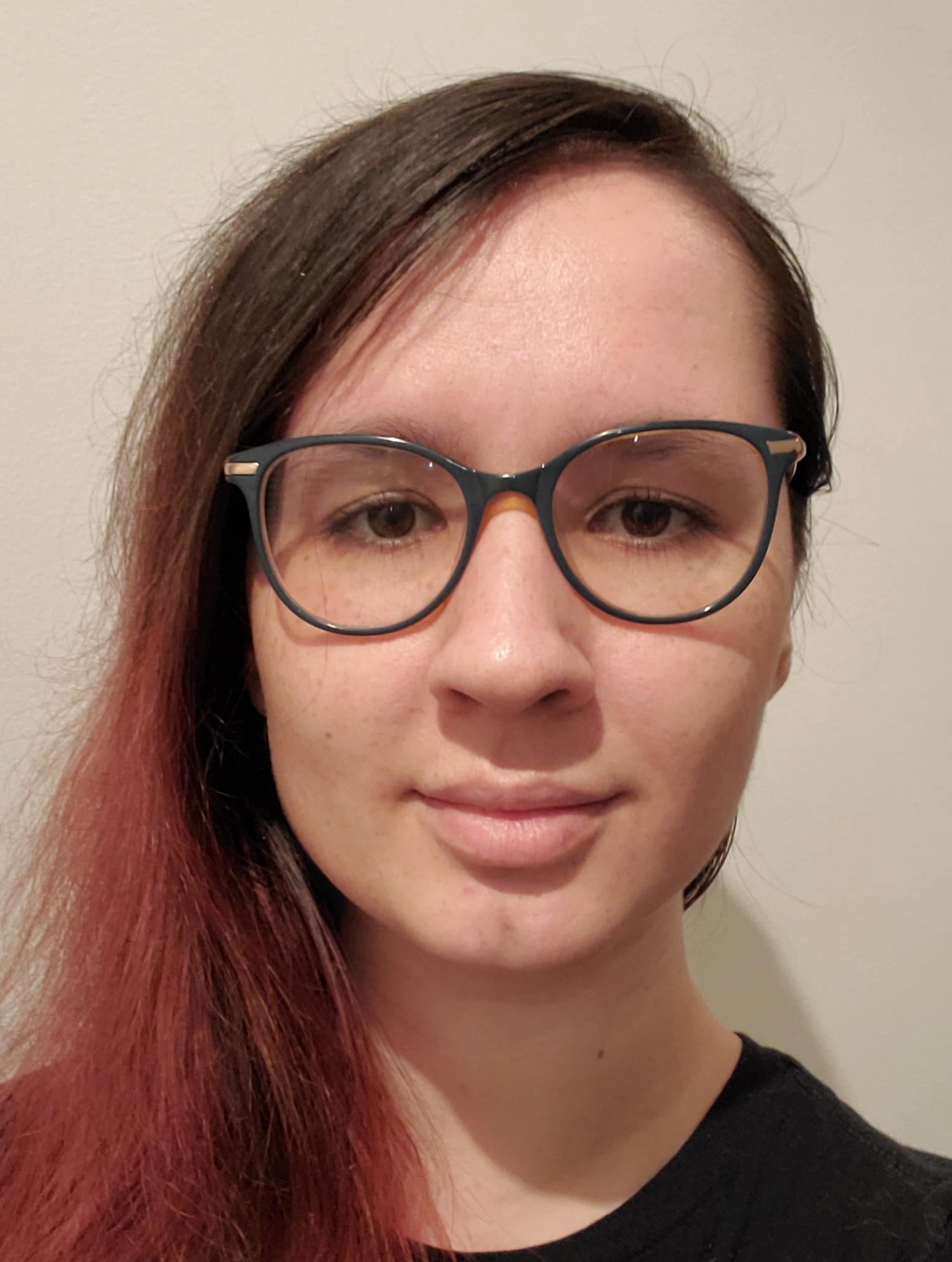}}]{Mhairi K MacLean}
received her M.Eng degree in biomedical engineering from the University of Glasgow, Glasgow, UK and her M.S in kinesiology from the University of Michigan, Ann Arbor, MI, USA. She received her PhD in biomedical engineering from the University of Florida, Gainesville, FL, USA. Mhairi is currently a postdoc in Hao Su’s laboratory at the City University of New York. Her current research interests include wearable robots and human biomechanics.
\end{IEEEbiography}

\begin{IEEEbiography}[{\includegraphics[width=1in,height=1.25in,clip,keepaspectratio]{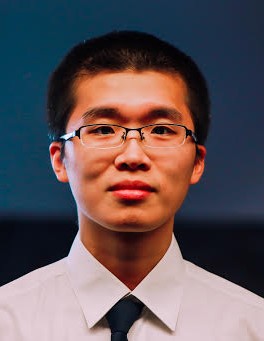}}]{Junxi Zhu}
received the B.S. from Shanghai Jiao Tong University and M.S. degree from the University of Maryland, College Park both in Mechanical Engineering. He is currently working towards a Ph.D. degree in Mechanical Engineering at the City University of New York, NY, USA under the supervision of Professor Hao Su. His current research interest includes controller design and implementation of the exoskeleton system.
\end{IEEEbiography}

\begin{IEEEbiography}[{\includegraphics[width=1in,height=1.25in,clip,keepaspectratio]{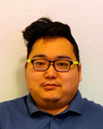}}]{Shuangyue Yu}
received the B.S. and M.S. degree from the Beijing University of Technology, Beijing, China. He is currently working towards the Ph.D. degree supervised by Dr. Hao Su at the Department of Mechanical Engineering at the City University of New York, New York, USA. His current research interests include lower-limb wearable robots.
\end{IEEEbiography}
\begin{IEEEbiography}[{\includegraphics[width=1in,height=1.25in,clip,keepaspectratio]{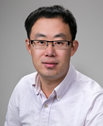}}]{Hao Su}
received the B.S. degree from the Harbin Institute of Technology, Harbin, China, the M.S. degree from the State University of New York University, Buffalo, NY, USA, and the Ph.D. degree from the Worcester Polytechnic Institute, Worcester, MA, USA. He is currently Irwin Zahn Endowed assistant professor in the Department of Mechanical Engineering at the City University of New York, New York, USA. His research interests include design and control of soft robots, wearable robots, and surgical robots. He was a recipient of the National Science Foundation CAREER Award. He currently serves as an associate editor of IEEE Robotics and Automation Letters (RAL), IEEE International Conference on Robotics and Automation (ICRA), and IEEE/RSJ International Conference on Intelligent Robots and Systems (IROS).
\end{IEEEbiography}





\vfill


\end{document}